\documentclass[10pt,twocolumn,letterpaper]{article}

\usepackage{iccv}
\usepackage{times}
\usepackage{epsfig}
\usepackage{graphicx}
\usepackage{amsmath}
\usepackage{amssymb}

\usepackage{algorithm}
\usepackage{algorithmic}
\usepackage{booktabs}
\usepackage{multirow}
\usepackage{float}
\usepackage{bbm}
\usepackage{relsize}
\newcommand\blfootnote[1]{%
  \begingroup
  \renewcommand\thefootnote{}\footnote{#1}%
  \addtocounter{footnote}{-1}%
  \endgroup
}

\usepackage[breaklinks=true,bookmarks=false]{hyperref}

\iccvfinalcopy % *** Uncomment this line for the final submission

\def\iccvPaperID{6625} % *** Enter the ICCV Paper ID here

\ificcvfinal\fi

\begin{document}

%%%%%%%%% TITLE
    \title{Variational Adversarial Active Learning}

\author{Samarth Sinha{\LARGE{\textbf{\textsuperscript{*}}}} \\
    University of Toronto\\
    % Institution1 address\\
    {\tt\small samarth.sinha@mail.utoronto.ca}
    % For a paper whose authors are all at the same institution,
    % omit the following lines up until the closing ``}''.
    % Additional authors and addresses can be added with ``\and'',
    % just like the second author.
    % To save space, use either the email address or home page, not both
    \and
    Sayna Ebrahimi{\LARGE{\textbf{\textsuperscript{*}}}}\\ % $^{~*}$ \\
    UC Berkeley\\
    {\tt\small sayna@eecs.berkeley.edu}
    \and
    Trevor Darrell\\
    UC Berkeley\\
    {\tt\small trevor@eecs.berkeley.edu}
}

\maketitle

% Remove page # from the first page of camera-ready.
\ificcvfinal\thispagestyle{empty}\fi

%%%%%%%%% ABSTRACT
    \begin{abstract}
    Active learning aims to develop label-efficient algorithms by sampling the most representative queries to be labeled by an oracle. We describe a pool-based semi-supervised active learning algorithm that implicitly learns this sampling mechanism in an adversarial manner. Unlike conventional active learning algorithms, our approach is \textit{task agnostic}, i.e., it does not depend on the performance of the task for which we are trying to acquire labeled data. Our method learns a latent space using a variational autoencoder (VAE) and an adversarial network trained to discriminate between unlabeled and labeled data. The mini-max game between the VAE and the adversarial network is played such that while the VAE tries to trick the adversarial network into predicting that all data points are from the labeled pool, the adversarial network learns how to discriminate between dissimilarities in the latent space. We extensively evaluate our method on various image classification and semantic segmentation benchmark datasets and establish a new state of the art on $\text{CIFAR10/100}$, $\text{Caltech-256}$, $\text{ImageNet}$, $\text{Cityscapes}$, and $\text{BDD100K}$. Our results demonstrate that our adversarial approach learns an effective low dimensional latent space in large-scale settings and provides for a computationally efficient sampling method.\blfootnote{{{{\textbf{\textsuperscript{*}}}}}Authors contributed equally, listed alphabetically.}\footnote{Our code and data are available at \url{https://github.com/sinhasam/vaal}.}
\end{abstract}

%%%%%%%%% BODY TEXT
    \section{Introduction}
\begin{figure}[t]
    \centering
    \includegraphics[width=0.5\textwidth]{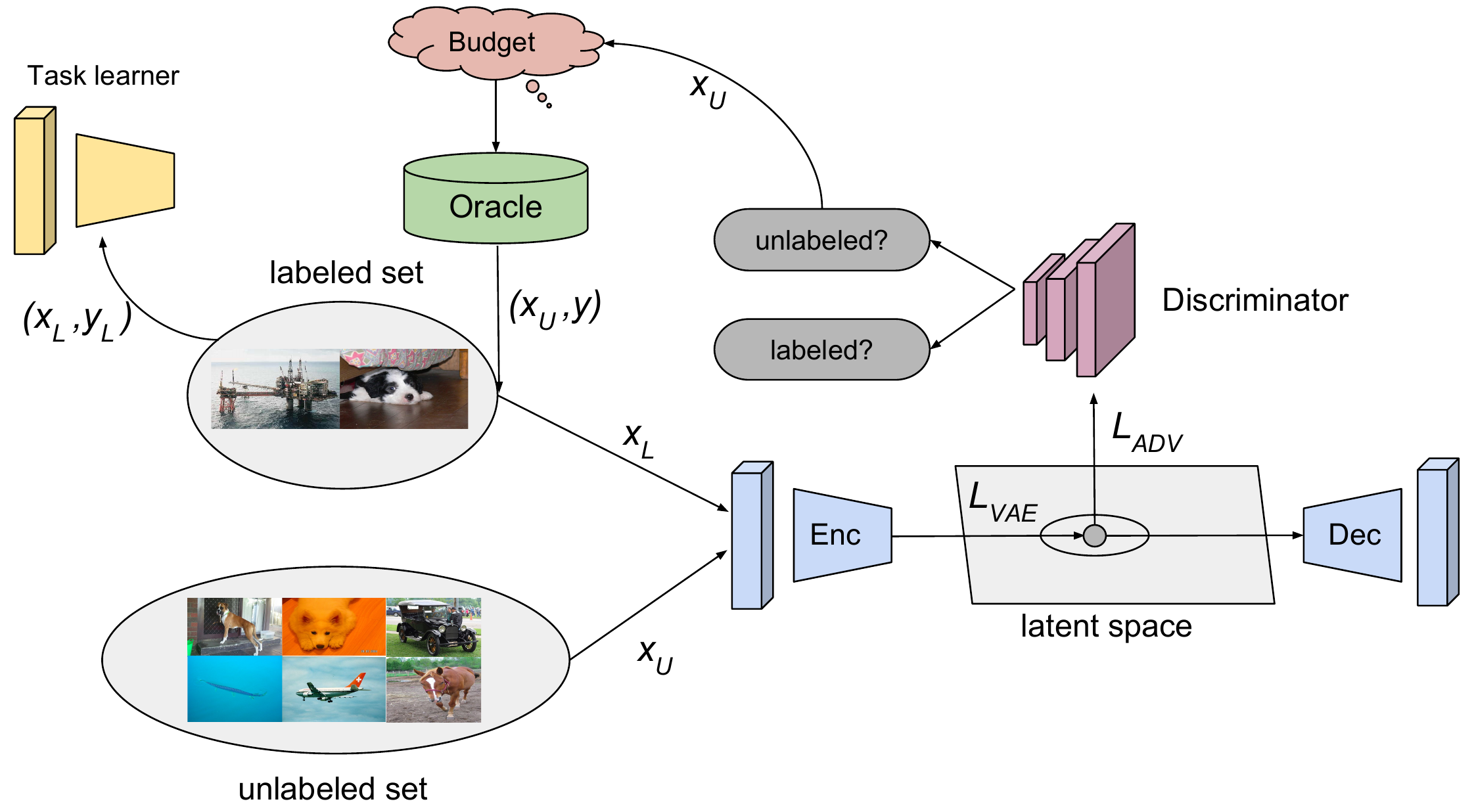}
    \caption{Our model (VAAL) learns the distribution of labeled data in a latent space using a VAE optimized using both reconstruction and adversarial losses.  A binary adversarial classifier (discriminator) predicts unlabeled examples and sends them to an oracle for annotations. The VAE is trained to fool the adversarial network to believe that all the examples are from the labeled data while the discriminator is trained to differentiate labeled from unlabeled samples. Sample selection is entirely separate from the main-stream task for which we are labeling data inputs, making our method to be \textit{task-agnostic}}
    \label{fig:teaser}
\end{figure}

The recent success of learning-based computer vision methods relies heavily on abundant annotated training examples, which may be prohibitively costly to label or impossible to obtain at large scale \cite{ebrahimi2017gradient}. In order to mitigate this drawback, active learning \cite{cohn} algorithms aim to incrementally select samples for annotation that result in high classification performance with low labeling cost. Active learning has been shown to require relatively fewer training instances when applied to computer vision tasks such as image classification \cite{sener2018coreset, li2013adaptive, gal2017deep, beluch2018ensemble} and semantic segmentation \cite{suggestiveannotation, qbcwithoutcoreset, gorriz2017costeffectivemelanoma}. 

This paper introduces a pool-based active learning strategy which learns a low dimensional latent space from labeled and unlabeled data using Variational Autoencoders (VAEs). VAEs have been well-studied and valued for both their generative properties as well as their ability to learn rich latent spaces. Our method, Variational Adversarial Active Learning (VAAL), selects instances for labeling from the unlabeled pool that are \textit{sufficiently} different in the latent space learned by the VAE to maximize the performance of the representation learned on the newly labeled data. Sample selection in our method is performed by an adversarial network which classifies which pool the instances belong to (labeled or unlabeled) and does not depend on the task or tasks for which are trying to collect labels. 

Our VAE learns a latent representation in which the sets of labeled and unlabeled data are mapped into a common embedding. We use an adversarial network in this space to correctly classify one from another. The VAE and the discriminator are framed as a two-player mini-max game, similar to GANs \cite{goodfellow2014generative} such that the VAE is trained to learn a feature space to \textit{trick} the adversarial network into predicting that all datapoints, from both the labeled and unlabeled sets, are from the labeled pool while the discriminator network learns how to discriminate between them. The strategy follows the intuition that once the active learner is trained, the probability associated with discriminator's predictions effectively estimates how representative each sample is from the pool that it has been deemed to be from. Therefore, instead of explicitly measuring uncertainty on the main task, we aim to choose points that would yield high uncertainty and thus are samples that are not well represented in the labeled set. We additionally consider oracles with different levels of labeling noise and demonstrate the robustness of our method to such noisy labels. In our experiments, we demonstrate superior performance on a variety of large scale image classification and segmentation datasets, and outperform current state of the art methods both in performance and computational cost.

%------------------------------------------------------------------
\section{Related Work}

\noindent \textbf{Active learning:} 
Current approaches can be categorized as query-acquiring (pool-based) or query-synthesizing methods. Query-synthesizing approaches use generative models to generate \textit{informative} samples  \cite{mahapatra2018efficient, adversarialsamplingactivelearning, zhu2017generative} whereas pool-based algorithms use different \textit{sampling strategies} to determine how to select the most \textit{informative} samples. Since our work lies in the latter line of research, we will mainly focus on previous work in this direction. 

Pool-based methods can be grouped into three major categories as follows: uncertainty-based methods \cite{gorriz2017costeffectivemelanoma, wang2017cost, beluch2018ensemble}, representation-based models \cite{sener2018coreset}, and their combination \cite{suggestiveannotation, preclustering}.
Pool-based methods have been theoretically proven to be effective and achieve better performance than the random sampling of points \cite{settles2014active, freund1997selective, gilad2006query}. Sampling strategies in pool-based algorithms have been built upon several methods, which are surveyed in \cite{activelearningsurvey}, such as information-theoretic methods \cite{mackay1992information}, ensembles methods \cite{mccallumzy1998employing, freund1997selective} and uncertainty heuristics such as distance to the decision boundary \cite{tong2001support} and conditional entropy \cite{li2013adaptive}. Uncertainty-based pool-based models are proposed in both Bayesian \cite{gal2017deep} and non-Bayesian frameworks. In the realm of Bayesian frameworks, probabilistic models such as Gaussian processes \cite{kapoor2007active, roy2001toward} or Bayesian neural networks \cite{ebrahimi2019uncertainty} are used to estimate uncertainty. Gal \& Gharamani \cite{gal2017deep, gal2016dropout}, also showed the relationship between uncertainty and dropout to estimate uncertainty in prediction in neural networks and applied it for active learning in small image datasets using shallow \cite{gal2016dropout} and deep \cite{gal2017deep} neural networks. In non-Bayesian classical active learning approaches, uncertainty heuristics such as distance from the decision boundary, highest entropy, and expected risk minimization have been widely investigated \cite{brinker2003incorporating, tong2001support, active_erm}. However, it was shown in \cite{sener2018coreset} that such classical techniques do not scale well to deep neural networks and large image datasets. Instead, they proposed to use Core-sets, where they minimize the Euclidean distance between the sampled points and the points that were not sampled in the feature space of the trained model \cite{sener2018coreset}. Using an ensemble of models to represent uncertainty was proposed by \cite{qbcwithoutcoreset, suggestiveannotation}, but \cite{diverseensembles} showed that using ensembles does not always yield high diversity in predictions which results in sampling redundant instances.  

Representation-based methods rely on selecting few examples by increasing \textit{diversity} in a given batch \cite{sener2018coreset, dutt2016active}. The Core-set technique was shown to be an effective representation learning method for large scale image classification tasks \cite{sener2018coreset} and was theoretically proven to work best when the number of classes is small. However, as the number of classes grows, it deteriorates in performance. Moreover, for high-dimensional data, using distance-based representation methods, like Core-set, appears to be ineffective because in high-dimensions \textit{p}-norms suffer from the curse of dimensionality which is referred to as the \textit{distance concentration phenomenon} in the computational learning literature \cite{franccois2008high}. We overcome this limitation by utilizing VAEs which have been shown to be effective in unsupervised and semi-supervised representation learning of high dimensional data \cite{vae, cvae}. 

Methods that aim to combine uncertainty and \textit{representativeness} use a two-step process to select the points with high uncertainty as of the most representative points in a batch. A hybrid framework combining uncertainty using conditional entropy and representation learning using information density was proposed in \cite{li2013adaptive} for classification tasks. A weakly supervised learning strategy was introduced in \cite{wang2017cost} that trains the model with pseudo labels obtained for instances with high \textit{confidence} in predictions. However, for a fixed performance goal, they often need to sample more instances per batch compared to other methods. Furthermore, in \cite{qbcwithoutcoreset} it was shown that having the representation step may not be necessary followed by suggesting an ensemble method that outperformed competitive approaches such as  \cite{suggestiveannotation} which uses uncertainty together with Core-sets. While we show that our model outperforms both \cite{qbcwithoutcoreset} and \cite{suggestiveannotation}, we argue that VAAL achieves this by learning the representation and uncertainty together such that they act in favor of each other while being independent from the main-stream task,  resulting in better active learning performance. 

\noindent \textbf{Variational autoencoders:} Autoencoders have long been used to effectively learn a feature space and representation \cite{bengio2013representation,schonfeld2019generalized}. A Variational AutoEncoder \cite{vae} is an example of a latent variable model that follows an encoder-decoder architecture of classical autoencoders which places a prior distribution on the feature space distribution and uses an Expected Lower Bound to optimize the learned posterior. %Another proposed models for representation learning include the Adversarial Autoencoder (AAE).
Adversarial autoencoders are a family of autoencoders which minimize the adversarial loss in the latent space between a sample from the prior and the posterior distribution \cite{makhzani2015adversarial}.  Prior work has investigated uncertainty modeling using a VAE for sequence generation in language applications \cite{thatijcaipaper}, 

\noindent \textbf{Active learning for semantic segmentation:}  Segmentation labeling is one of the most expensive annotations to collect. Active learning in the literature has been broadly investigated for labeling medical images as it is one of the most prevailing applications of AL where only human experts with sophisticated knowledge are capable of providing labels and therefore, improving this process would reduce a lot of time and effort for them. Suggestive Annotation (SA) \cite{suggestiveannotation} uses uncertainty obtained from an ensemble of models trained on the labeled data and Core-sets for choosing representative data points in a two-step strategy. \cite{qbcwithoutcoreset} also proposed an active learning algorithm for image segmentation using an ensemble of models, but they empirically showed their proposed information-theoretic heuristic for uncertainty is equal in performance to SA, without using Core-sets.  \cite{gorriz2017costeffectivemelanoma} extended the work by \cite{gal2017deep} and proposed using Monte-Carlo dropout masks on the unlabeled images using a trained model and calculating the uncertainty on the predicted labels of the unlabeled images. Some active learning strategies developed for image classification can also be used for semantic segmentation. Core-sets and max-entropy strategies can both be used for active learning in semantic segmentation \cite{sener2018coreset, brinker2003incorporating}.

\noindent \textbf{Adversarial learning:} Adversarial learning has been used for different problems such as generative models \cite{goodfellow2014generative}, representation learning \cite{makhzani2015adversarial, mescheder2017adversarial}%, factorvae, wae}
, domain adaptation \cite{adda, cycada}, deep learning robustness and security \cite{madry2017towards, tramer2017ensemble} etc. The use of an adversarial network enables the model to train in a fully-differentiable by adjusting to solving the \textit{mini-max} optimization problem  \cite{goodfellow2014generative}. The adversarial network used in the feature space has been extensively researched in the representation learning and domain adaptation literature to efficiently learn a useful feature space for the task \cite{makhzani2015adversarial, factorvae, wae, adda, cycada}. 

\section{Adversarial Learning of Variational Auto-encoders for Active Learning}

Let ($x_L, y_L$) be a sample pair belonging to the pool of labeled data ($X_L, Y_L$). $X_U$ denotes a much larger pool of samples ($x_U$) which are not yet labeled. The goal of the active learner is to train the most label-efficient model by iteratively querying a fixed sampling \textit{budget}, $b$ number of the most informative samples from the unlabeled pool ($x_U \sim X_U$), using an acquisition function to be annotated by the oracle such that the expected loss is minimized. 

\subsection{Transductive representation learning.} 
We use a $\beta$-variational autoencoder for representation learning in which the encoder learns a low dimensional space for the underlying distribution using a Gaussian prior and the decoder reconstructs the input data. In order to capture the features that are missing in the representation learned on the labeled pool, we can benefit from using the unlabeled data and perform transductive learning. The objective function of the $\beta$-VAE is minimizing the variational lower bound on the marginal likelihood of a given sample formulated as
{\small{
\begin{eqnarray}\label{eq:tr}
\mathcal{L}_{\mathrm{VAE}}^{trd}=&\mathbb{E}[\log{p}_\theta(x_L | z_L )] - \beta~\text{D}_{\mathrm{KL}}(q_{\phi}(z_L|x_L) || p(z)) 
&  \nonumber\\ & +\mathbb{E}[\log{p}_\theta(x_{U} | z_{U} )] - \beta~\text{D}_{\mathrm{KL}}(q_{\phi}(z_{U}|x_{U}) || p(z)) 
\end{eqnarray}
}}
\noindent where $q_\phi$ and $p_\theta$ are the encooder and decoder parameterized by  $\phi$ and $\theta$, respectively. $p(z)$ is the prior chosen as a unit Gaussian, and $\beta$ is the Lagrangian parameter for the optimization problem. The reparameterization trick is used for proper calculation of the gradients \cite{vae}.

\subsection{Adversarial representation learning }

The representation learned by the VAE is a mixture of the latent features associated with both labeled and unlabeled data. An ideal active learning agent is assumed to have a perfect sampling strategy that is capable of sending the most \textit{informative} unlabeled data to the oracle. Most of the sampling strategies rely on the model's uncertainty, i.e, the more uncertain the model is on the prediction, the more informative that specific unlabeled data must be. However, this introduces vulnerability to the outliers. In contrast we train an adversarial network for our sampling strategy to learn how to distinguish between the encoded features in the latent space. This adversarial network is analogous to discriminators in GANs where their role is to discriminate between fake and real images created by the generator. In VAAL, the adversarial network is trained to map the latent representation of $z_L \cup z_U$ to a binary label which is $1$ if the sample belongs to $X_L$ and is $0$, otherwise. The key to our approach is that the VAE and the adversarial network are learned together in an adversarial fashion. While the VAE maps the labeled and unlabeled data into the same latent space with similar probability distribution $q_{\phi}(z_{L}|x_{L})$ and $q_{\phi}(z_{U}|x_{U})$, it fools the discriminator to classify all the inputs as labeled. On the other hand, the discriminator attempts to effectively estimate the probability that the data comes from the unlabeled data. We can formulate the objective function for the adversarial role of the VAE as a binary cross-entropy loss as below
%\begin{equation}\label{eq:vae-adv}
%\mathcal{L}_\mathrm{VAE}^{adv} = \mathcal{L}_{\mathrm{BCE}}(q_{\phi}(z_L|x_L), \mathbbm{1}) + \mathcal{L}_{\mathrm{BCE}}(q_{\phi}(z_U|x_U)), \mathbbm{1})
%\end{equation}
{\small{
\begin{equation}\label{eq:vae-adv}
\mathcal{L}_{\mathrm{VAE}}^{adv} = -\mathbb{E}[\log(D(q_{\phi}(z_L | x_L)))] - \mathbb{E}[\log(D(q_{\phi}(z_U | x_U)))]
\end{equation}
}}
%where $\mathcal{L}_{\mathrm{BCE}}$ is simply a binary cross-entropy cost function. 
The objective function to train the discriminator is also given as below
%\begin{equation}\label{eq:dis}
%\mathcal{L}_\mathrm{D} = \mathcal{L}_{\mathrm{BCE}}(q_{\phi}(z_L|x_L)), \mathbbm{1}) + \mathcal{L}_{\mathrm{BCE}}(q_{\phi}(z_U|x_U)), \mathbb{O})
%\end{equation}
{\small{
\begin{equation}\label{eq:dis}
\mathcal{L}_D = -\mathbb{E}[\log(D(q_{\phi}(z_L | x_L)))] - \mathbb{E}[\log(1 - D(q_{\phi}(z_U | x_U)))] 
\end{equation}
}}
By combining Eq. (\ref{eq:tr}) and Eq. (\ref{eq:vae-adv}) we obtain the full objective function for the VAE in VAAL as below
\begin{equation}\label{eq:vae}
\mathcal{L}_{\mathrm{VAE}} = \lambda_1 \mathcal{L}_{\mathrm{VAE}}^{trd} + \lambda_2 \mathcal{L}_\mathrm{VAE}^{adv}
\end{equation}
where $\lambda_{1}$ and $\lambda_2$ are hyperparameters that determine the effect of each component to learn an effective variational adversarial representation.

The task module denoted as $T$ in Fig. (\ref{fig:teaser}), learns the task for which the active learner is being trained. $T$ is trained separately from the active learner as they do not depend on each other at any step. We report results below on image classification and semantic segmentation tasks, using VGG16 \cite{vgg} and dilated residual network (DRN) architecture \cite{drn} with an unweighted cross-entropy cost function. Our full algorithm is shown in Alg. \ref{alg:vaal}.

\subsection{Sampling strategies and noisy-oracles}   
The labels provided by the oracles might vary in how  \textit{accurate} they are  depending on the quality of available human resources. For instance, medical images annotated by expert humans are assumed to be more accurate than crowd-sourced data collected by non-expert humans and/or available information on the cloud. 
We consider two types of oracles: an ideal oracle which always provides correct labels for the active learner, and a noisy oracle which non-adversarially provides erroneous labels for some specific classes. This might occur due to similarities across some classes causing ambiguity for the labeler. In order to present this oracle realistically, we have applied a targeted noise on visually similar classes. 
The sampling strategy in VAAL is shown in Alg. (\ref{alg:sampling}). We use the probability associated with the discriminator's predictions as a score to collect $b$ number of samples in every batch predicted as ``unlabeled'' with the lowest confidence to be sent to the oracle. Note that the closer the probability is to zero, the more likely it is that it comes from the unlabeled pool. The key idea to our approach is that instead of relying on the performance of the training alforithm on the main-stream task, which suffers from being inaccurate specially in the beginning, we select samples based on the likelihood of their \textit{representativeness} with respect to other samples which discriminator thinks belong to the unlabeled pool.

\begin{algorithm}[t]
    \caption{Variational Adversarial Active Learning} \label{alg:vaal}
    \begin{algorithmic}[1]
        \renewcommand{\algorithmicrequire}{\textbf{Input:}}
        \renewcommand{\algorithmicensure}{\textbf{Output:}}
        \REQUIRE Labeled pool ($X_{L}, Y_{L})$, Unlabeled pool $(X_{U})$, Initialized models for $\theta_{T}$, $\theta_{VAE}$, and $\theta_{D}$
        \REQUIRE Hyperparameters: epochs, $\lambda_1$, $\lambda_2$,  $\alpha_1$,  $\alpha_2$,  $\alpha_3$
        \FOR {$e = 1$ \text{to epochs}}
        \STATE sample $(x_L, y_L) \sim (X_L, Y_L$)
        \STATE sample $x_U \sim X_U$
        \STATE Compute $\mathcal{L}_\mathrm{VAE}^{trd}$ by using Eq. \ref{eq:tr} 
        \vspace{3pt}
        \STATE Compute $\mathcal{L}_\mathrm{VAE}^{adv}$ by using Eq. \ref{eq:vae-adv} 
        \vspace{3pt}     
        \STATE $\mathcal{L}_{\mathrm{VAE}} \gets \lambda_1 \mathcal{L}_{\mathrm{VAE}}^{trd} + \lambda_2 \mathcal{L}_\mathrm{VAE}^{adv}$ 
        \vspace{1pt}
        \STATE Update VAE by descending stochastic gradients: 
        \STATE $\theta'_{VAE} \gets \theta_{VAE} - \alpha_1 \nabla \mathcal{L}_{\mathrm{VAE}} $
%        \STATE {\footnotesize{$\mathcal{L}_\mathrm{D} \gets \mathcal{L}_{\mathrm{BCE}}(q_{\phi}(z_L|x_L)), \mathbbm{1}) + \mathcal{L}_{\mathrm{BCE}}(q_{\phi}(z_U|x_U)), \mathbb{O})$}}
        \STATE Compute $\mathcal{L}_\mathrm{D}$ by using Eq. \ref{eq:dis} 
        \vspace{1pt}
        \STATE Update $D$ by descending its stochastic gradient:
        \STATE $\theta'_{D} \gets \theta_{D} - \alpha_2 \nabla \mathcal{L}_\mathrm{D} $
        
        \vspace{3pt}
        \STATE Train and update $T$:
        \STATE $\theta'_{T} \gets \theta_T - \alpha_3 \nabla \mathcal{L}_{\mathrm{T}} $
        \ENDFOR
        
        \RETURN Trained $\theta_{T}, \theta_{VAE}, \theta_{D}$ 
    \end{algorithmic} 
\end{algorithm}

\begin{algorithm}[t]
    \caption{Sampling Strategy in VAAL} \label{alg:sampling}
    \begin{algorithmic}[1]
        \renewcommand{\algorithmicrequire}{\textbf{Input:}}
        \renewcommand{\algorithmicensure}{\textbf{Output:}}
        \REQUIRE $b, X_{L}, X_{U}$
        \ENSURE $X_{L}, X_{U}$
        %    \STATE $\mu_{x_{U}} \leftarrow \theta_{VAE}(X_{U})$
        %    \STATE $y_{U} \leftarrow \theta_{adv}(\mu_{x_{U}})$
        \STATE Select samples ($X_s$) with $\min_b\{\theta_{D}(z_U)\}$ %and feed them to oracle
        %    \STATE $x_{sample} \leftarrow \min_k\{y_{U}\}$
        %\STATE $y_{o} \leftarrow \mathcal{ORACLE}(X_{s})$
        %\STATE $X_{L} \leftarrow X_{L} \cup (X_{s}, y_{o})$
        \STATE $Y_{o} \leftarrow \mathcal{ORACLE}(X_{s})$
        \STATE $(X_L,Y_L) \leftarrow (X_L,Y_L) \cup (X_s,Y_o)$
        \STATE $X_U \leftarrow X_U - {X_{s}}$
        \RETURN $X_{L}, X_{U}$
    \end{algorithmic} 
\end{algorithm}

\section{Experiments}

We begin our experiments with an initial labeled pool with $10\%$ of the training set labeled. The budget size per batch is equal to $5\%$ of the training dataset. The pool of unlabeled data contains the rest of the training set from which samples are selected to be annotated by the oracle. Once labeled, they will be added to the initial training set and training is repeated on the new training set. We assume the oracle is \textit{ideal} unless stated otherwise. \footnote{Data and code required to reproduce all plots are provided at \url{https://github.com/sinhasam/vaal/blob/master/plots/plots.ipynb}.}

\noindent \textbf{Datasets.} We have evaluated VAAL on two common vision tasks. For image classification we have used $\text{CIFAR10}$ \cite{cifar} and $\text{CIFAR100}$ \cite{cifar} both with $60$K images of size $32\times32$, and $\text{Caltech-256}$ \cite{caltech256} which has $30607$ images of size $224\times224$ including $256$ object categories. For a better understanding of the scalability of VAAL we have also experimented with $\text{ImageNet}$ \cite{imagenet} with more than $1.2$M images of $1000$ classes. For semantic segmentation, we evaluate our method on $\text{BDD100K}$ \cite{bdd100k} and $\text{Cityscapes}$ \cite{cityscapes} datasets both of which have $19$ classes. $\text{BDD100K}$ is a diverse driving video dataset with $10$K images with full-frame instance segmentation annotations collected from distinct locations in the United State. $\text{Cityscapes}$ is also another large scale driving video dataset containing $3475$ frames with instance segmentation annotations recorded in street scenes from $50$ different cities in Europe. The statistics of these datasets are summarized in Table \ref{tab:dataset_table} in the appendix.

\begin{figure*}[t]
    \centering
    \includegraphics[width=0.47\textwidth]{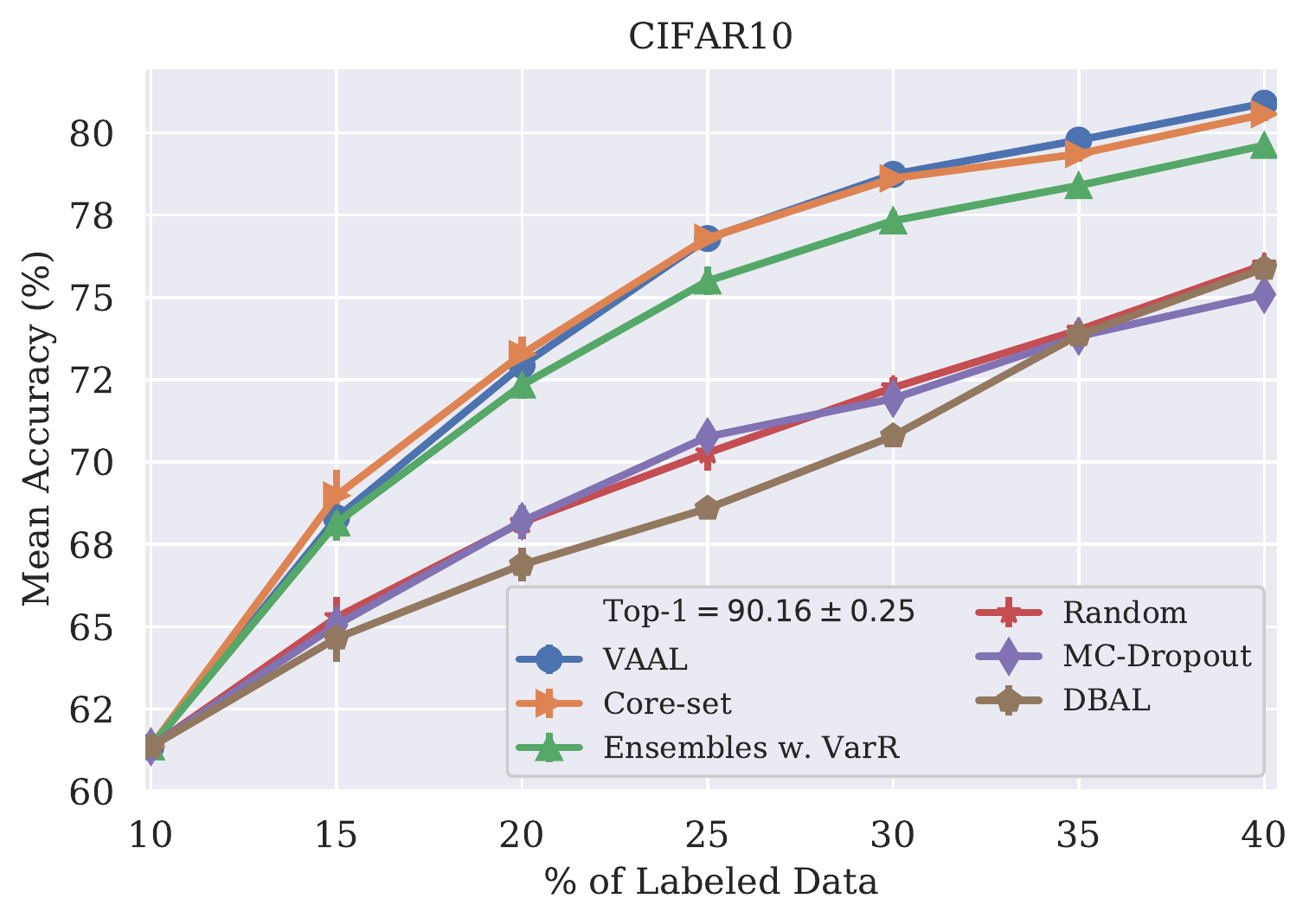}
    \includegraphics[width=0.47\textwidth]{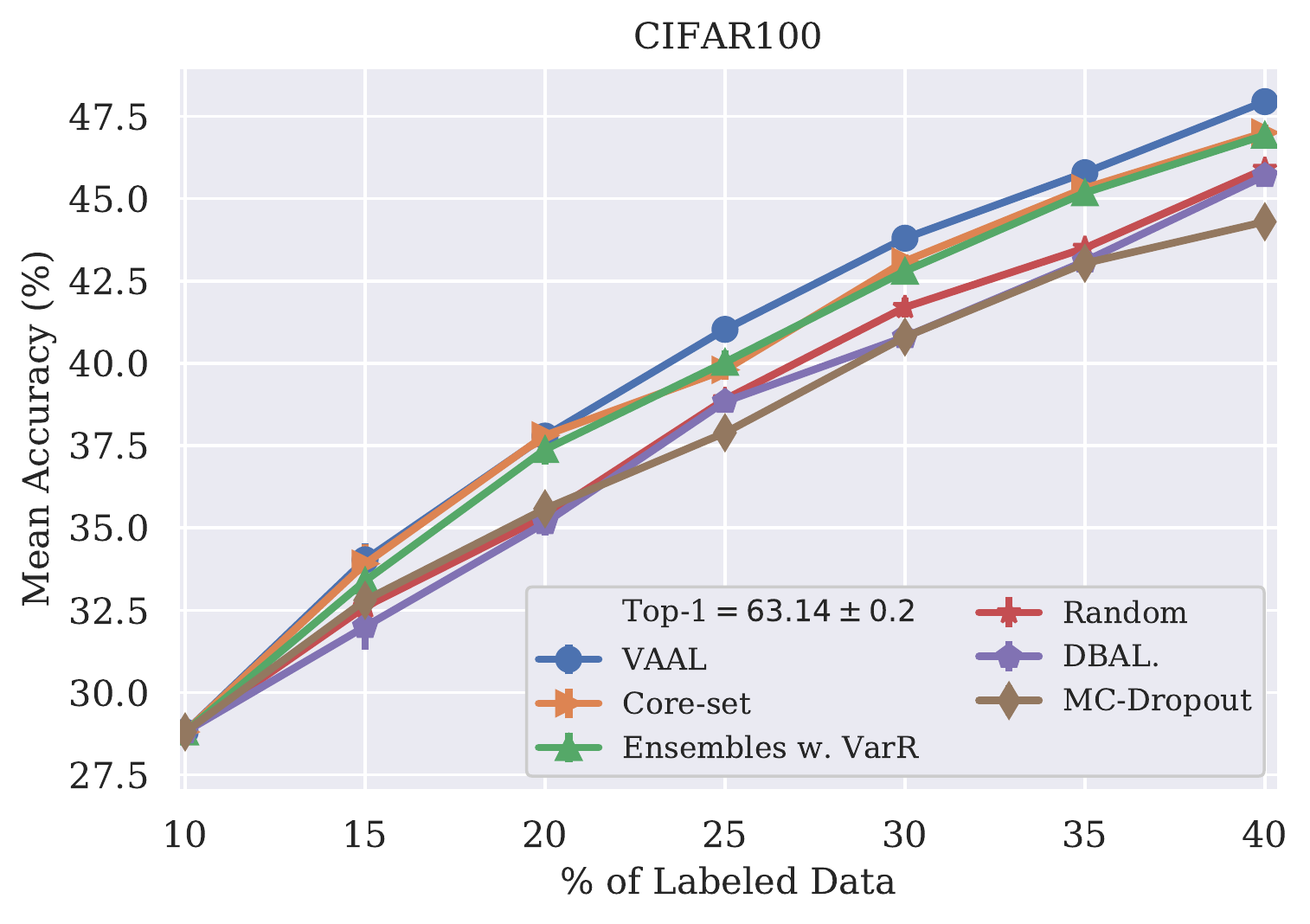} 
    \includegraphics[width=0.47\textwidth]{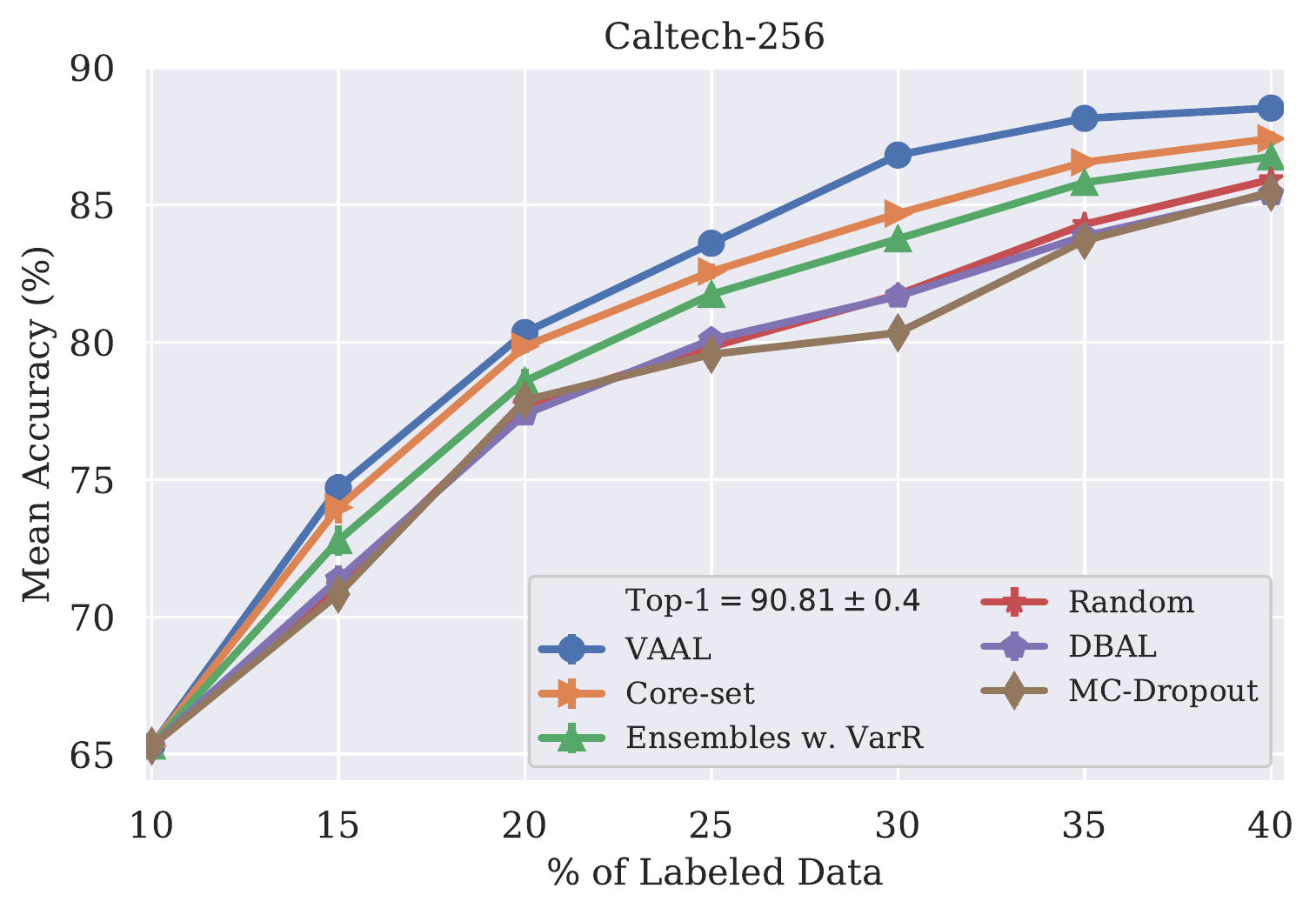}
    \includegraphics[width=0.47\textwidth]{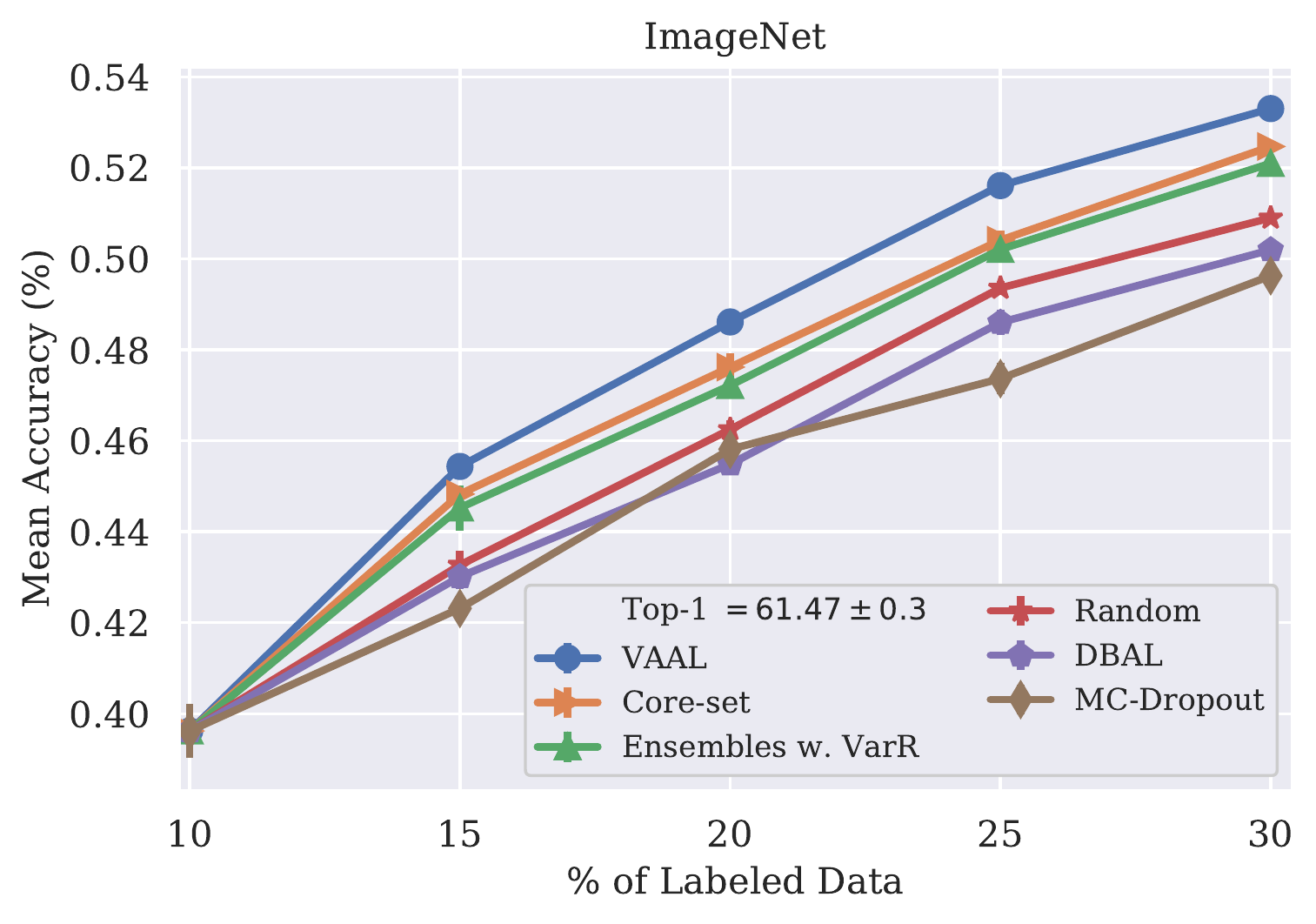}
    \caption{VAAL performance on classification tasks using $\text{CIFAR10}$, $\text{CIFAR100}$, $\text{Caltech-256}$, and $\text{ImageNet}$ compared to Core-set \cite{sener2018coreset}, Ensembles w. VarR \cite{beluch2018ensemble}, $\text{MC-Dropout}$ \cite{gal2016dropout}, DBAL \cite{gal2017deep}, and Random Sampling. Best visible in color. Data and code required to reproduce are provided in our code repository}
    \label{fig:classification}
\end{figure*}

\noindent \textbf{Performance measurement.}
We evaluate the performance of VAAL in image classification and segmentation by measuring the accuracy and mean IoU, respectively achieved by $T$ trained with $10\%$, $15\%$, $20\%$, $25\%$, $30\%$, $35\%$, $40\%$ of the total training set as it becomes available with labels provided by the oracle. Results for all our  experiments, except for $\text{ImageNet}$, are averaged over $5$ runs. $\text{ImageNet}$ results however, are obtained by averaging over $2$ repetitions
using $10\%$, $15\%$, $20\%$, $25\%$, $30\%$ of the training data.

\subsection{VAAL on image classification benchmarks} \label{sec:classification} 

\noindent \textbf{Baselines.} We compare our results using VAAL for image classification against various approaches including Core-set \cite{sener2018coreset}, Monte-Carlo Dropout \cite{gal2016dropout}, and Ensembles using Variation Ratios (Ensembles w. VarR) \cite{beluch2018ensemble, freeman1965elementary}. We also show the performance of deep Bayesian AL (DBAL) by following \cite{gal2017deep} and perform sampling using their proposed max-entropy scheme to measure uncertainty \cite{shannon1948mathematical}. We also show the results using \textit{random sampling} in which samples are uniformly sampled at random from the unlabeled pool. This method still serves as a competitive baseline in active learning. Moreover, we use the mean accuracy achieved on the entire dataset as an upper bound which does not adhere to the active learning scenario. 

\noindent \textbf{Implementation details.} We used random horizontal flips for data augmentation. The architecture used in the task module for image classification is VGG16 \cite{vgg} with Xavier initialization \cite{xavier_init} and $\beta$-VAE has the same architecture as the Wasserstein autoencoder \cite{wae} with latent dimensionality given in Table \ref{tab:hyperparams} in the appendix. The discriminator is a $5$-layer multilayer perceptron (MLP) and Adam \cite{kingma2014adam} is used as the optimizer for all these three modules with an equal learning rate of $5\times10^{-4}$ and batch size of $64$. However, for $\text{ImageNet}$, learning rate varies across the modules such that the task learner has a learning rate of $1\times10^{-1}$ while the VAE and the discriminator have a learning rate of $5\times10^{-3}$. Training continues for $100$ epochs in $\text{ImageNet}$ and for $100$ epochs in all other datasets. The budget size for classification experiments is chosen to be $5\%$ of the full training set, which is equivalent to $2500$, $2500$, $1530$, and $64060$ for $\text{CIFAR10}$, $\text{CIFAR100}$, $\text{Caltech-256}$, and $\text{ImageNet}$, respectively in VAAL and all other baselines. A complete list of hyperparameters used in our model are found through a grid search and are tabulated in Table \ref{tab:hyperparams} in the appendix.

\noindent \textbf{VAAL performance $\text{CIFAR10/100}$ and $\text{Caltech-256}$.} Figure \ref{fig:classification} shows performance of VAAL compared to prior works. On $\text{CIFAR10}$, our method achieves mean accuracy of $80.9\%$ by using $40\%$ of the data whereas using the entire dataset yields accuracy of $90.16\%$, denoted as Top-1 accuracy in Fig. \ref{fig:classification}. %whereas this value is $75.99\%$ for random sampling. 
Comparing the mean accuracy values for data ratios above $15\%$ shows that VAAL evidently outperforms random sampling, DBAL, and $\text{MC-Dropout}$ while beating Ensembles by a smaller margin and becoming on-par with Core-set. On $\text{CIFAR100}$, VAAL remains competitive with Ensembles w. VarR and Core-set, and outperforms all other baselines. The maximum achievable mean accuracy is $63.14\%$ on $\text{CIFAR100}$ using $100\%$ of the data while VAAL achieves $47.95\% $ by only using $40\%$ of it. Moreover, for data ratios above $20\%$ of labeled data, VAAL consistently requires $\sim2.5\%$ less number of labels compared to Core-set or Ensembles w. VarR  in order to achieve the same accuracy, which is equal to $1250$ labels. On $\text{Caltech-256}$, which has real images of object categories, VAAL consistently outperforms all baselines by an average margin of $1.78\%$ from random sampling and $1.01\%$ from the most competitive baseline, Core-set. DBAL method performs nearly identical to random sampling while $\text{MC-Dropout}$ yields lower accuracies than random sampling. By looking at the number of labels required to reach a fixed performance, for instance, $83.6\%$, VAAL needs $25\%$ of data ($7651$ images) to be labeled whereas this number is approximately $9200$ and $9500$ for Core-set and Ensemble w. VarR, respectively. Random sampling, DBAL, and $\text{MC-Dropout}$ all need more than $12200$ images. 

As can be seen in Fig. \ref{fig:classification}, VAAL outperforms Core-set with higher margins as the number of classes increases from $10$ to $100$ to $256$. The theoretical analysis shown in \cite{sener2018coreset} confirms that Core-set is more effective when fewer classes are present due to the negative impact of high dimensionality on \textit{p}-norms in the Core-set method. %This is more apparent in our experiments on the $\text{ImageNet}$ dataset which has $1000$ classes described in the following subsection. 

\noindent \textbf{VAAL performance on $\text{ImageNet}$.}
$\text{ImageNet}$ \cite{imagenet} is a challenging large scale dataset which we use to show scalability of our approach.
Fig. \ref{fig:classification} shows that we improve the state-of-the-art by $100\%$ increase in the gap between the accuracy achieved by the previous state-of-the-art methods (Core-set and Ensemble) and random sampling. As can be seen in Fig. \ref{fig:classification}, this improvement can be also viewed in the number of samples required to achieve a specific accuracy. For instance, the accuracy of $48.61\%$ is achieved by VAAL using $256$K number of images whereas Core-set and Ensembles w. VarR should be provided with almost $32$K more labeled images to obtain the same performance. Random sampling remains as a competitive baseline as both DBAL and MC-Dropout perform below that.

\subsection{VAAL on image segmentation benchmarks}
\noindent \textbf{Baselines.} 
We evaluate VAAL against state-of-the-art AL approaches for image segmentation including Core-set \cite{sener2018coreset}, $\text{MC-Dropout}$ \cite{gorriz2017costeffectivemelanoma}, Query-By-Committee (QBC) \cite{qbcwithoutcoreset}, and suggestive annotation (SA)\cite{suggestiveannotation}. SA is a hybrid ensemble method that uses bootstrapping for uncertainty estimation \cite{bootstrap} and core-set for measuring \textit{representativeness}.

\noindent\textbf{Implementation details.} Similar to the image classification setup, we used random horizontal flips for data augmentation. The $\beta$-VAE is a Wasserstein autoencoder \cite{wae}, and the discriminator is also a $5$-layer MLP. The architecture used in the task module for image segmentation is DRN \cite{drn} and Adam with a learning rate of $5\times10^{-4}$ is chosen as the optimizer for all three modules. The batch size is set as $8$ and training stops after $50$ epochs in both datasets. The budget size used in VAAL and all baselines is set as $400$ and $150$ for $\text{BDD100K}$ and $\text{Cityscapes}$, respectively. All hyperparameteres are shown in Table \ref{tab:hyperparams} in the appendix 

\begin{figure}[t]
    \centering
    \includegraphics[scale=0.48]{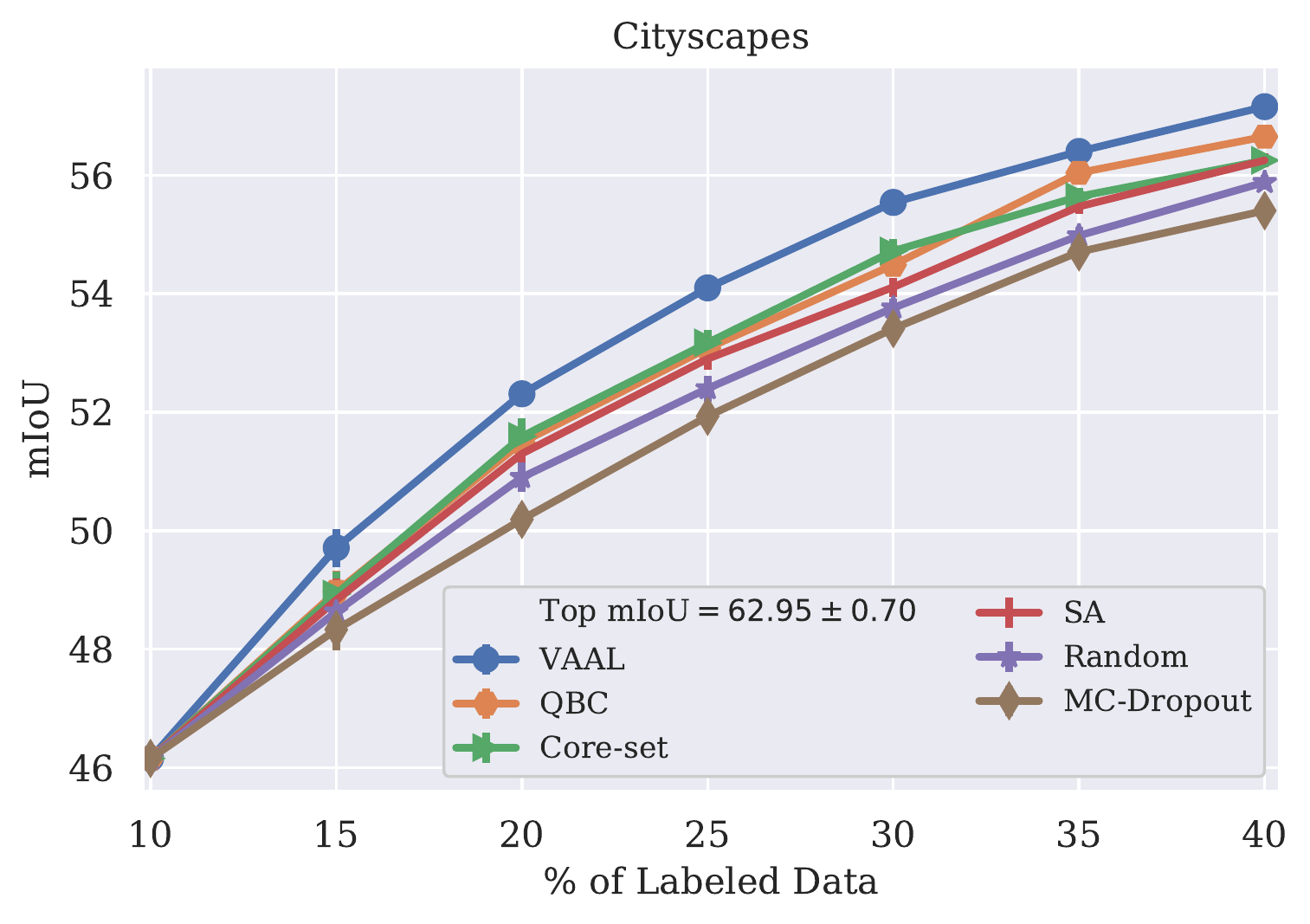}
    \includegraphics[scale=0.48]{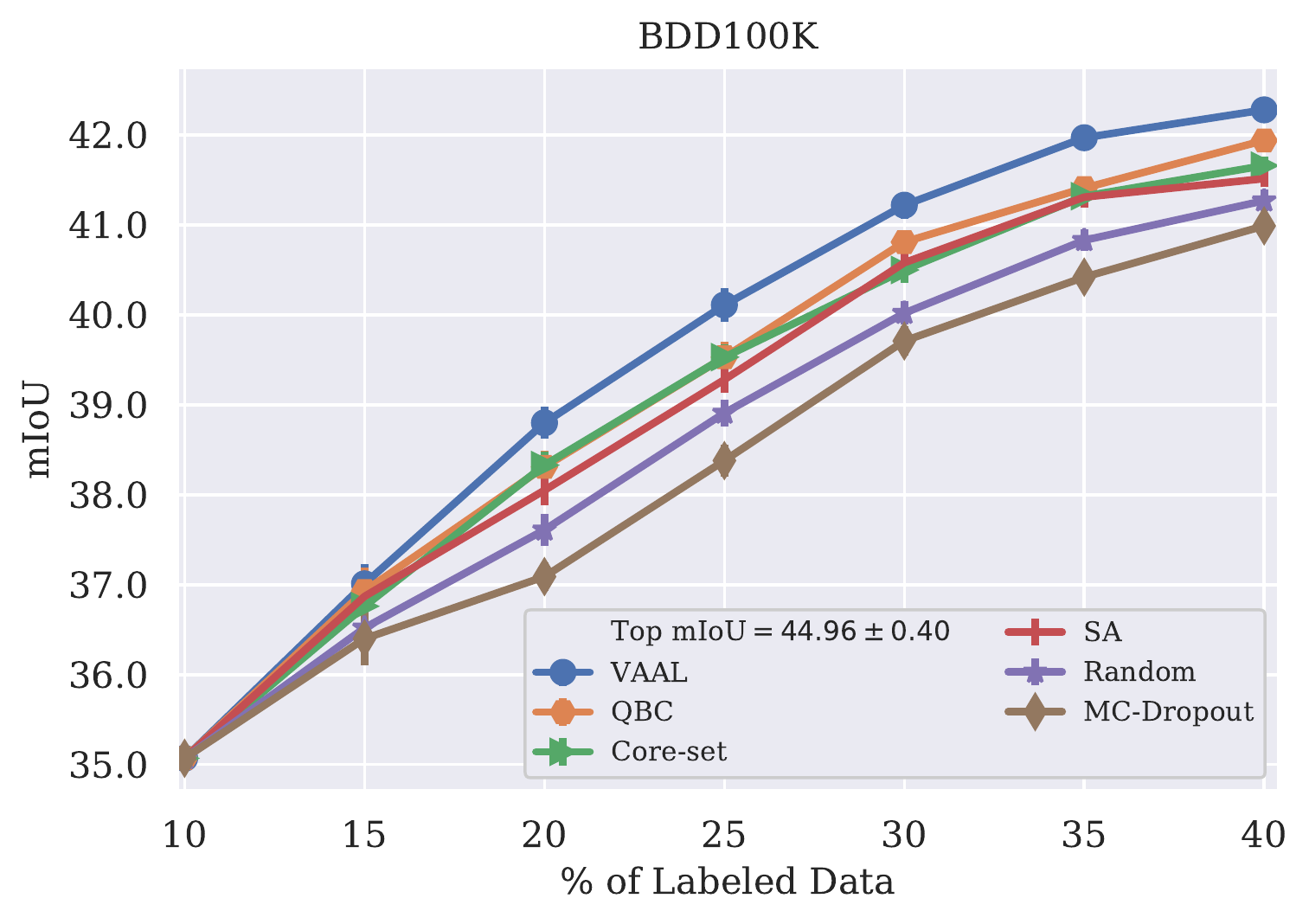}
    \caption{VAAL performance on segmentation tasks using $\text{Cityscapes}$ and $\text{BDD100K}$ compared to QBC \cite{qbcwithoutcoreset}, Core-set \cite{sener2018coreset}, $\text{MC-Dropout}$ \cite{gal2016dropout}, and Random Sampling. Data and code required to reproduce are provided in our code repository}
    %     \vskip -0.05in
    \label{fig:seg}
\end{figure}

\noindent\textbf{VAAL performance on $\text{Cityscapes}$ and $\text{BDD100K}$.} Figure \ref{fig:seg} demonstrates our results on the driving datasets compared with four other baselines as well as the reference random sampling. As we also observed in section \ref{sec:classification} Core-set performs better with fewer number of classes in image classification tasks \cite{sener2018coreset} . However, the large gap between VAAL and Core-set, despite only having $19$ classes, suggests that Core-set and Ensemble-based methods (QBC in here) suffer from high dimensionality in the inputs ($688\times688$ as opposed to thumbnail $32\times32$ images used in $\text{CIFAR10/100}$). QBC and Core-set, and SA (Core-set + QBC) perform nearly identical, while $\text{MC-Dropout}$ remains less effective than random sampling. VAAL consistently demonstrate significantly better performance by achieving the highest mean IoU on both $\text{Cityscapes}$ and $\text{BDD100K}$ across different labeled data ratios. VAAL is able to achieve $\%$mIoU of $57.2$ and $42.3$ using only $40\%$ labeled data while the maximum mIoU we obtained using $100\%$ of these datasetes is $62.95$ and $44.95$ on $\text{Cityscapes}$ and $\text{BDD100K}$, respectively. In terms of required labels by each method, on $\text{Cityscapes}$ VAAL needs $743$ annotations to reach $54.1\%$ of mIoU whereas QBC, Core-set, SA, random sampling, $\text{MC-Dropout}$ demand nearly $800$, $890$, $910$, $960$, and $1041$ labels, respectively. Similarly on $\text{BDD100K}$ in order to reach $41\%$ of mIoU, other baselines need $5\%-10\%$ more annotations than VAAL requires only $30\%$. Considering the difficulties in full frame instance segmentation, VAAL is able to effectively reduce the required time and effort for such dense annotations.

\section{Analyzing VAAL in Detail}
In this section, we take a deeper look into our model by first performing ablation and then evaluating the effect of possible biases and noise on its performance. Sensitivity of VAAL to budget size is also explored in \ref{sec:budget}.

\subsection{Ablation study}

\begin{figure}[t]
    \begin{center}
        \includegraphics[scale=0.5]{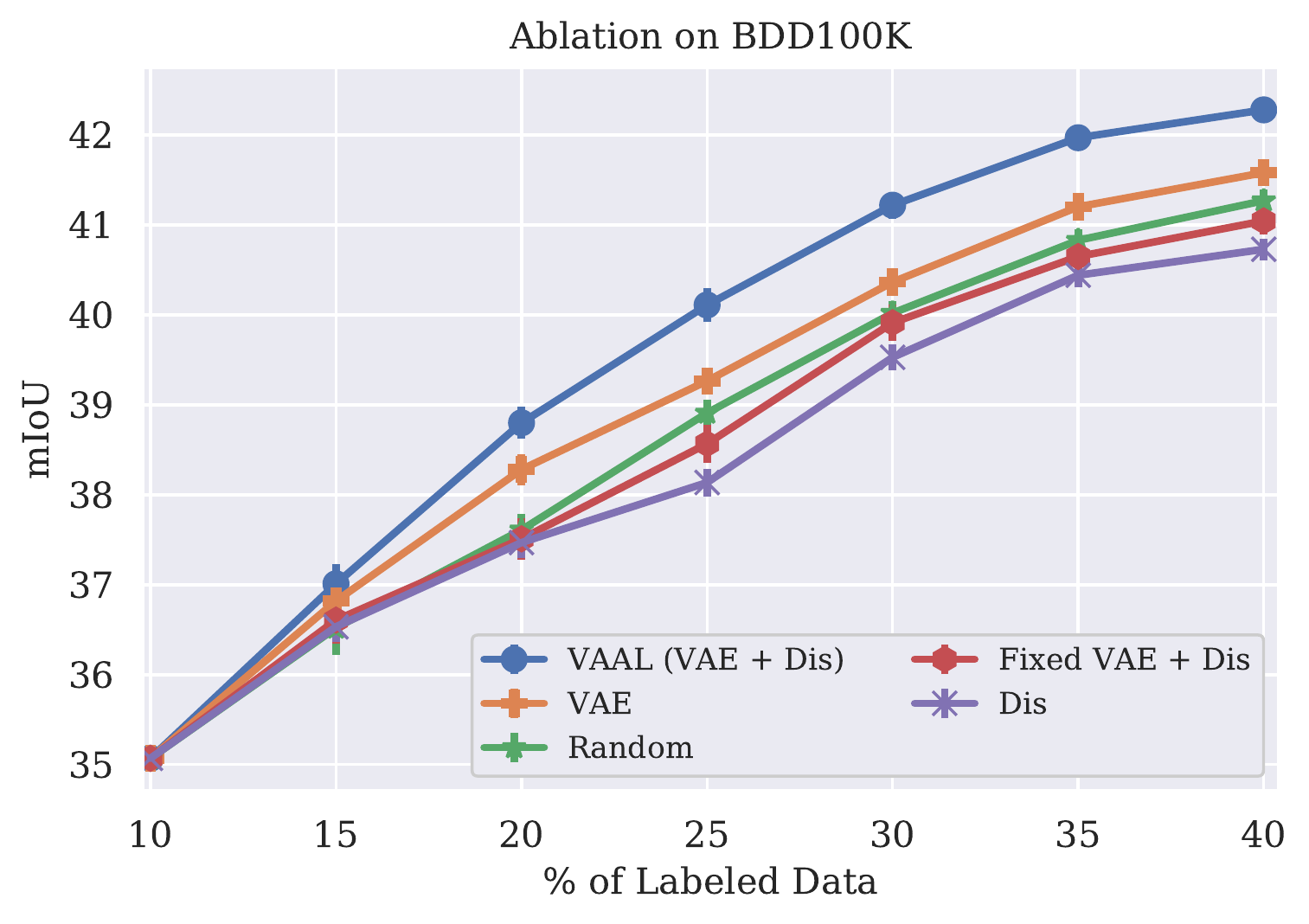}
    \end{center}
    \caption{Ablation results on analyzing the effect of the VAE and the discriminator denoted as $Dis$ here. Data and code required to reproduce are provided in our code repository}
    \label{fig:ablation}
\end{figure}

Figure \ref{fig:ablation} presents our ablation study to inspect the contribution of the key modules in VAAL including the VAE, and the discriminator ($D$). We perform ablation on the segmentation task which is more challenging than classification and we use $\text{BDD100K}$ as it is larger than $\text{Cityscapes}$. The variants of ablations we consider are: 1) eliminating VAE, 2) Frozen VAE with D, 3) eliminating $D$. In the first ablation, we explore the role of the VAE as the representation learner by having only a discriminator trained on the image space to discriminate between labeled and unlabeled pool. As shown in Fig. \ref{fig:ablation}, this setting results in the discriminator to only memorize the data and yields the lowest performance. Also, it reveals the key role of the VAE in not only learning a rich latent space but also playing an effective mini-max game with the discriminator to avoid overfitting. In the second ablation scenario, we add a VAE to the previous setting to encode-decode a lower dimensional space for training $D$. However, here we avoid training the VAE and hence merely explore its role as an autoencoder. This setting performs better than having only the $D$ trained in a high dimensional space, but yet performs similar or worse than random sampling suggesting that discriminator failed at learning \textit{representativeness} of the samples in the unlabeled pool. In the last ablation, we explore the role of the discriminator by training only a VAE that uses 2-Wasserstein distance from the cluster-centroid of the labeled dataset as a heuristic to explicitly measure uncertainty. For a multivariate isotropic Gaussian distribution, the closed-form solution for the 2-Wasserstein distance between two probability distributions \cite{givens1984class} can be written as 
\begin{equation}
\mathrm{W_{ij}} = \big[||\mu_i - \mu_j||_2^2 + ||\Sigma_i^{\frac{1}{2}} - \Sigma_j^{\frac{1}{2}}||_\mathcal{F}^2 \big]^\frac{1}{2}
\end{equation}
where $||.||_\mathcal{F}$ represents the Frobenius norm and $\mu_i$, $\Sigma_i$ denote the $\mu$, $\Sigma$ predicted by the encoder and $\mu_j$, $\Sigma_j$ are the mean and variance for the normal distribution over the labeled data from which the latent variable $z$ is generated. In this setting, we see an improvement over random sampling which shows the effect of explicitly measuring the uncertainty in the learned latent space. However, VAAL appears to outperform all these scenarios by implicitly learning the uncertainty over the adversarial game between the discriminator and the VAE.

\subsection{VAAL's Robustness}

\begin{figure*}[t]
    \begin{center}
        \includegraphics[scale=0.42]{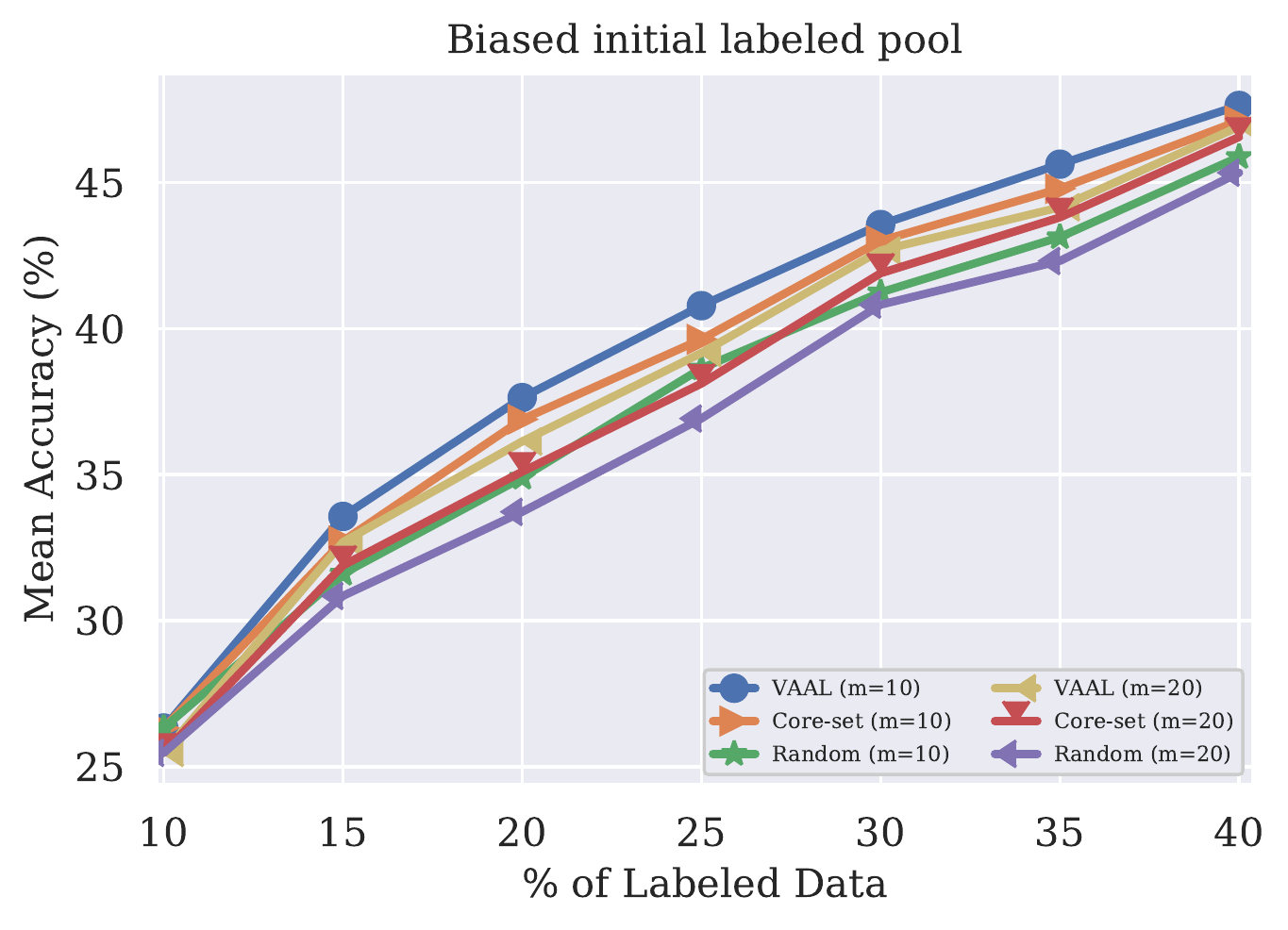}
        \includegraphics[scale=0.42]{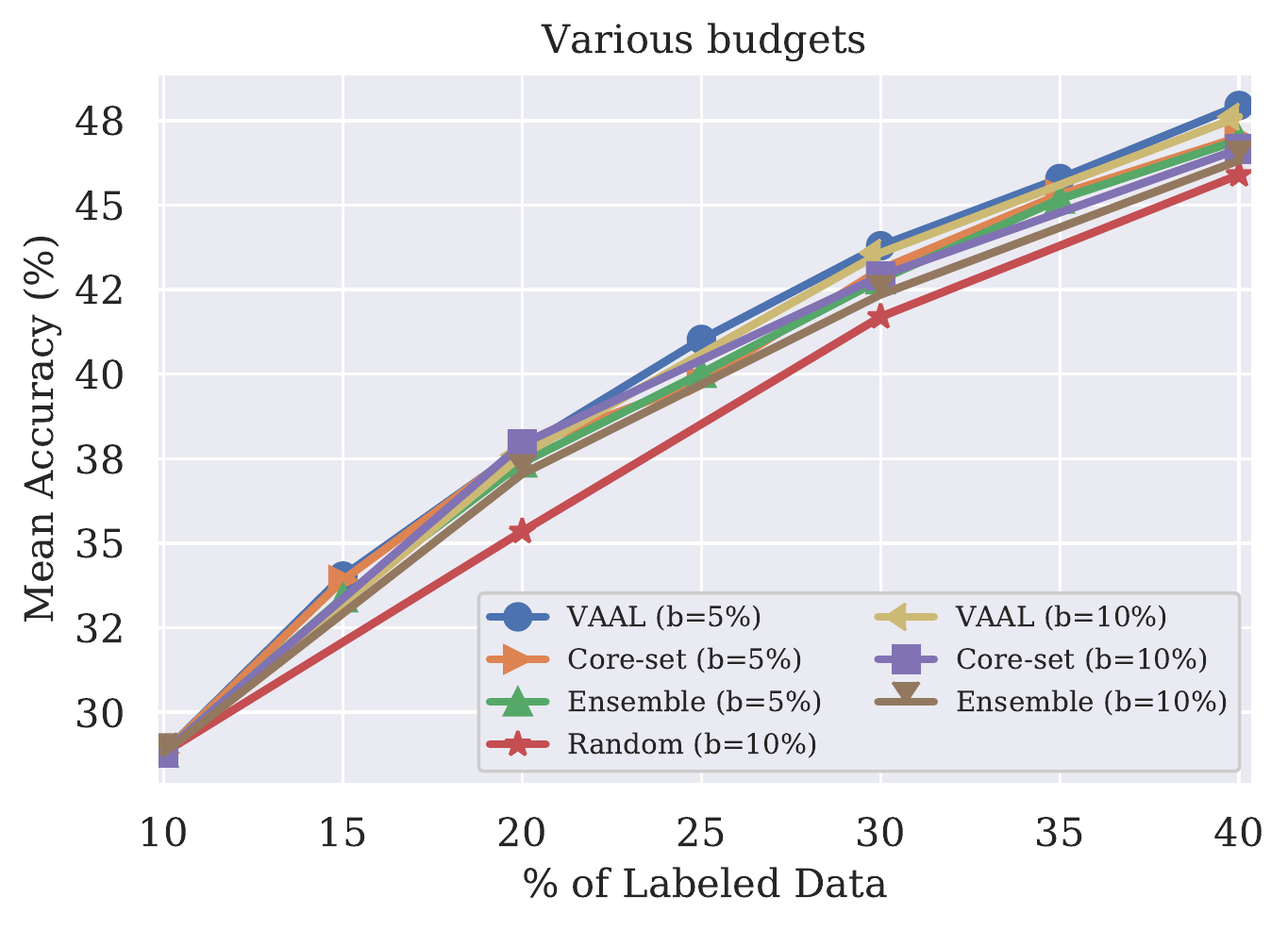}
        \includegraphics[scale=0.36]{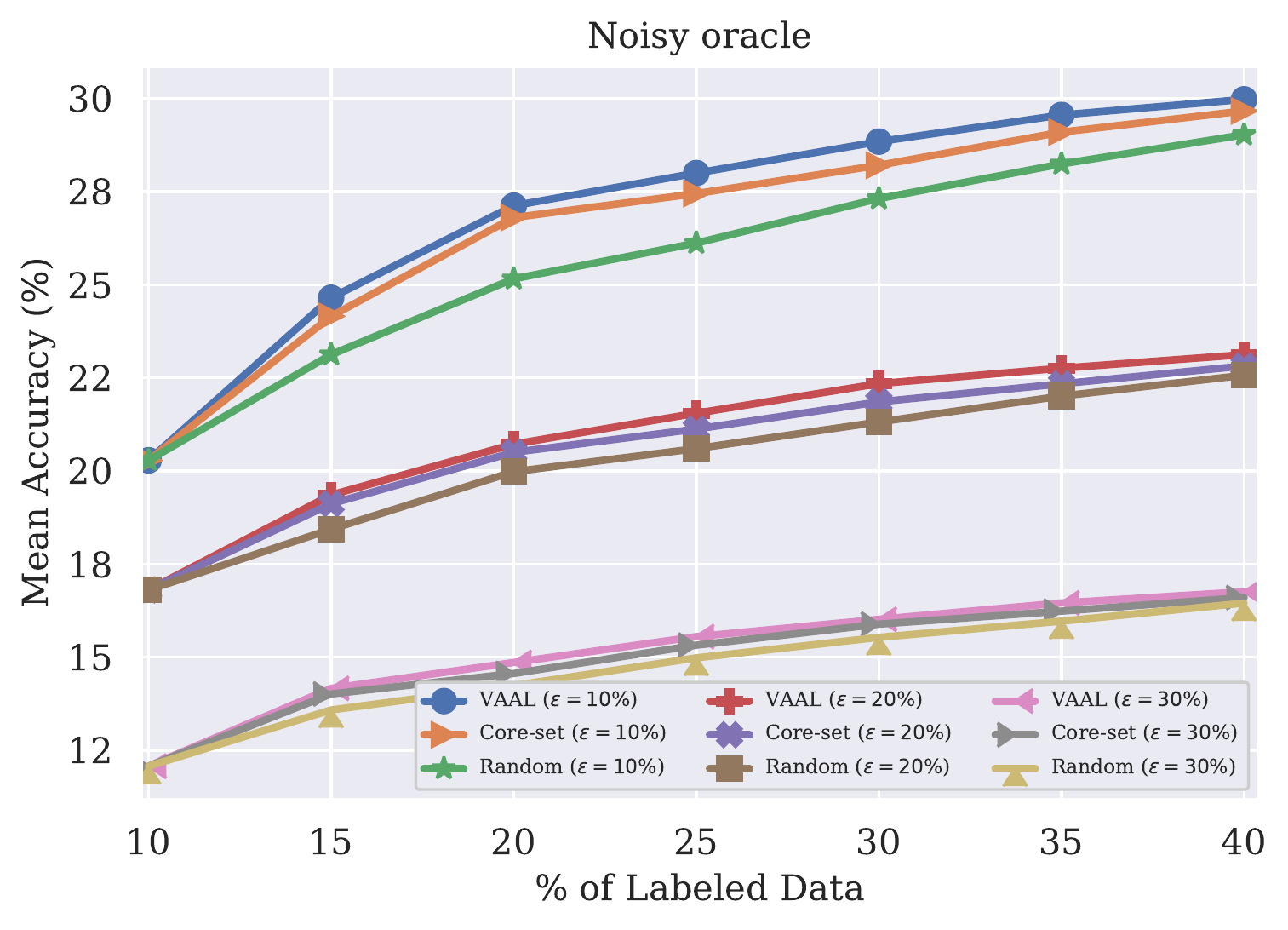}
    \end{center}
    \caption{Analyzing robustness of VAAL to noisy labels, budget size, and biased initial labeled pool using $\text{CIFAR100}$. Best viewed in color. Data and code required to reproduce are provided in our code repository}
    \label{fig:robustness}
\end{figure*}

\noindent \textbf{Effect of biased initial labels in VAAL.}
We investigate here how bias in the initial labeled pool affect VAAL's performance as well as other baselines on $\text{CIFAR100}$ dataset. Intuitively, bias can affect the training such that it causes the initially labeled samples to be not representative of the underlying data distribution by being inadequate to cover most of the regions in the latent space. We model a possible form of bias in the labeled pool by not providing labels for $m$ chosen classes at random and we compare it to the case where samples are randomly selected from all classes. We exclude the data for $m=10$ and $m=20$ classes at random in the initial labeled pool to explore how it affects the performance of the model. Figure \ref{fig:robustness} shows for $m=10$ and $m=20$, VAAL is superior to Core-set and random sampling in selecting informative samples from the classes that were underrepresented in the initial labeled set. We also observe that VAAL with $m=20$ missing classes performs nearly identical to Core-Set and significantly better than random sampling where each has half number of missing classes.

\noindent \textbf{Effect of budget size on performance. }\label{sec:budget}
Figure \ref{fig:robustness} illustrates the effect of the budget size on our model compared to the most competitive baselines on $\text{CIFAR100}$. We repeated our experiments in section \ref{sec:classification} for a lower budget size of $b=5\%$. We observed that VAAL outperforms Core-Set and Ensemble w. VarR, as well as random sampling, on both budget sizes of $b=5\%$ and $b=10\%$. Core-set comes at the second best method followed by Ensemble in Fig \ref{fig:robustness}. We note that $b=5\%$ for all methods, including VAAL, has a slightly better performance compared to when $b=10\%$ which is expected to happen because a larger sampled batch results in adding redundant samples instead of more informative ones. 

\noindent \textbf{Noisy vs. ideal oracle in VAAL. }\label{sec:noise} In this analysis we investigate the performance of VAAL in the presence of noisy data caused by an inaccurate oracle. We assume the erroneous labels are due to the ambiguity between some classes and are not adversarial attacks. We model the noise as targeted noise on specific classes that are \textit{meaningful} to be mislabeled by a human labeler. We used $\text{CIFAR100}$ for this analysis because of its  hierarchical structure in which $100$ classes in $\text{CIFAR100}$ are grouped into $20$ super-classes. Each image comes with a \textit{fine} label (the class to which it belongs) and a \textit{coarse} label (the super-class to which it belongs). We randomly change the ground truth labels for $10\%$, $20\%$ and $30\%$ of the training set to have an incorrect label within the same super-class. Figure \ref{fig:robustness} shows how a noisy oracle effects the performance of VAAL, Core-set, and random sampling. Because both Core-set and VAAL do not depend on the task learner, we see that the relative performance is comparable to the ideal oracle presented in Section \ref{sec:classification}. Intuitively, as the percentage of noisy labels increases, all of the active learning strategies converge to random sampling.

\noindent \textbf{Choice of the network architecture in $T$. }\label{sec:arch}
In order to assure VAAL is insensitive to the VGG16 architecture used in our classification experiments, we also used ResNet18 \cite{resnet} in VAAL and the most competitive baseline (Core-set). Figure \ref{fig:arch} in the appendix shows the choice of the architecture does not affect the performance gap between VAAL and Core-set. 

\begin{table}[t]\footnotesize
    \centering
    \begin{tabular}{|c|c|}
        \hline
        Method & Time (sec)  \\
        \hline
        \hline
        $\text{MC-Dropout}$ \cite{gal2016dropout} & $81.05$ \\
        \hline        
        Core-set \cite{sener2018coreset} & $75.33$ \\
        \hline
        Ensembles w. VarR \cite{beluch2018ensemble} & $20.48$ \\
        \hline
        DBAL. \cite{gal2017deep} & $10.95$  \\
        \hline
        \hline
        \textbf{VAAL (ours)} & {$\mathbf{10.59}$} \\
        \hline
    \end{tabular}
    \vskip 0.05in
    \caption{Time taken to sample, for one sampling iteration, from the unlabeled pool on $\text{CIFAR10}$ dataset. For a fair comparison we use the same PyTorch data-loader across VAAL and baselines.}
    \vskip -0.2in
    \label{tab:time_comparison}
\end{table}

\subsection{Sampling time analysis}
The sampling strategy of an active learner has to select samples in a time-efficient manner. In other words, it should be as close as possible to random sampling, considering the fact that random sampling is still an effective baseline. Table \ref{tab:time_comparison} shows our comparison for VAAL and all our baselines on $\text{CIFAR10}$ using a single NVIDIA TITAN Xp. Table \ref{tab:time_comparison} shows the time needed to sample a fixed budget of images from the unlabeled pool for all the methods. $\text{MC-Dropout}$ performs multiple forward passes to measure the uncertainty from $10$ dropout masks which explains why it appears to be very slow in sample selection. Core-set and Ensembles w. VarR, are the most competitive baselines to VAAL in terms of their achieved mean accuracy. However, in sampling time, VAAL takes $10.59$ seconds while Core-set requires $75.33$ sec and Ensembles w. VarR needs $20.48$ sec. DBAL \cite{gal2017deep} is on-par in sampling time with VAAL, however, DBAL is outperformed in accuracy by all other methods including random sampling which can sample in only a few milliseconds. 
The significant difference between Core-set and VAAL is due to the fact that Core-set needs to solve an optimization problem for sample selection as opposed to VAAL which only needs to perform inference on the discriminator and rank its output probabilities. The Ensembles w. VarR method uses $5$ models to measure the uncertainty resulting in better computational efficiency but it does not yet perform as fast as VAAL.

\section{Conclusion}
In this paper we proposed a new batch mode task-agnostic active learning algorithm, VAAL, that learns a latent representation on both labeled and unlabeled data in an adversarial game between a VAE and a discriminator, and implicitly learns the uncertainty for the samples deemed to be from the unlabeled pool. We demonstrate state-of-the-art results, both in terms of accuracy and sampling time, on small and large-scale image classification ($\text{CIFAR10}$, $\text{CIFAR100}$, $\text{Caltech-256}$, $\text{ImageNet}$) and segmentation datasets ($\text{Cityscapes}$, $\text{BDD100K}$). We further showed that VAAL is robust to noisy labels and biased initial labeled data, and it performs consistently well, given different oracle budgets.

%------------------------------------------------------------------------

{\small
\bibliographystyle{ieee_fullname}
\bibliography{ref}
}

%------------------------------------------------------------------------
%------------------------------------------------------------------------
%------------------------------------------------------------------------
%------------------------------------------------------------------------
%------------------------------------------------------------------------
%
\clearpage
\onecolumn
\section*{\centering SUPPLEMENTARY MATERIAL \hfil} 
\vskip -0.1in
\section*{A. Datasets}
\vskip -0.1in
Table \ref{tab:dataset_table} shows a summary of the datasets utilized in our work along with their size and number of classes and budget size.
\vskip -0.1in

\begin{table}[ht]\footnotesize
    \vskip 0.15in
    \centering
    \begin{tabular}{|c|c|c|c|c|c|c|}
        \hline
        & & & & Initially & & \\
        Dataset & $\#$Classes & Train + Val & Test & Labeled & Budget & Image Size\\
        \hline
        \hline
        $\text{CIFAR10}$ \cite{cifar}& $10$ & $45000+5000$  & $10000$  & $5000$ & $2500$ & $32\times32$\\
        \hline
        $\text{CIFAR100}$ \cite{cifar}& $100$ & $45000+5000$ & $10000$ & $5000$ & $2500$ & $32\times32$ \\
        \hline
        $\text{Caltech-256}$ \cite{caltech256}& $256$ & $27607+3000$ & $2560$ & $3060$ &  $1530$ & $224\times224$\\
        \hline
        $\text{ImageNet}$ \cite{imagenet} & $1000$ & $1153047+128120$ & $50000$ & $128120$ &  $64060$ & $224\times224$ \\
        \hline
        $\text{BDD100K}$ \cite{bdd100k} & $19$ & $7000 + 1000$ & $2000$ & $800$ &  $400$ & $688\times688$\\
        \hline
        $\text{Cityscapes}$ \cite{cityscapes} & $19$ & $2675+300$ & $500$ & $300$ & $150$ & $688\times688$\\
        \hline
    \end{tabular}
    \vskip 0.09in    
    \caption{A summary of the datasets used in our experiments. $\text{CIFAR10}$, $\text{CIFAR100}$, $\text{Caltech-256}$ and $\text{ImageNet}$ are datasets used for image classification, while $\text{BDD100K}$ and $\text{Cityscapes}$ are large scale segmentation datasets. The budget for each dataset is the number of images that can be sampled at each training iteration.}
    \label{tab:dataset_table}
\end{table}    

\section*{B. Hyperparameter Selection}
\vskip -0.1in
\noindent Table \ref{tab:hyperparams} shows the hyperparameters found for our models through a grid search.
\vskip -0.1in    

\begin{table}[ht]\footnotesize
    \vskip 0.15in
    \centering
    \begin{tabular}{|c|c|c|c|c|c|c|c|c|c|}
        \hline
        Experiment & $d$ & $\alpha_1$ & $\alpha_2$ & $\alpha_3$ &  $\lambda_1$  & $\lambda_2$ & $\beta$ & batch size & epochs \\
        \hline\hline
        $\text{CIFAR10}$ & $32$ & $5\times10^{-4}$ & $5\times10^{-4}$ & $5\times10^{-4}$ & $1$ & $1$ & $1$ &$64$ & $100$ \\
        \hline
        $\text{CIFAR100}$ &  $32$ & $5\times10^{-4}$ & $5\times10^{-4}$ & $5\times10^{-4}$ & $1$ & $1$ & $1$ &$64$ & $100$ \\
        \hline
        $\text{Caltech-256}$ & $64$ &$5\times10^{-4}$ & $5\times10^{-4}$ & $5\times10^{-4}$ & $1$ & $10$ & $1$ &$64$ & $100$ \\
        \hline
        $\text{ImageNet}$ &  $64$  &$10^{-1}$ & $10^{-3}$ & $10^{-3}$ & $1$ & $10$ & $1$ &$64$ & $100$ \\
        \hline
        $\text{BDD100K}$ & $128$  &$10^{-3}$ & $10^{-3}$ & $10^{-3}$ & $1$ & $25$ & $1$ &$8$ & $100$ \\
        \hline
        $\text{Cityscapes}$ &  $128$  & $10^{-3}$ & $10^{-3}$ & $10^{-3}$ & $1$ & $25$  &  $1$ & $8$ & $100$ \\
        \hline
    \end{tabular}
    \vskip 0.08in 
    \caption{Hyperparameters used in our experiments for VAAL. $d$ is the latent space dimension of VAE. $\alpha_1$, $\alpha_2$, and $\alpha_3$ are learning rates for VAE, discriminator ($D$), and task module ($T$), respectively. $\lambda_1$ and $\lambda_2$ are the regularization parameters for transductive and adversarial terms used in Eq. (\ref{eq:vae}). $\beta$ is the Lagrangian parameter in Eq. (\ref{eq:tr}).}
    \label{tab:hyperparams}
\end{table}

\begin{figure}[H]
    \begin{center}
        \includegraphics[scale=0.6]{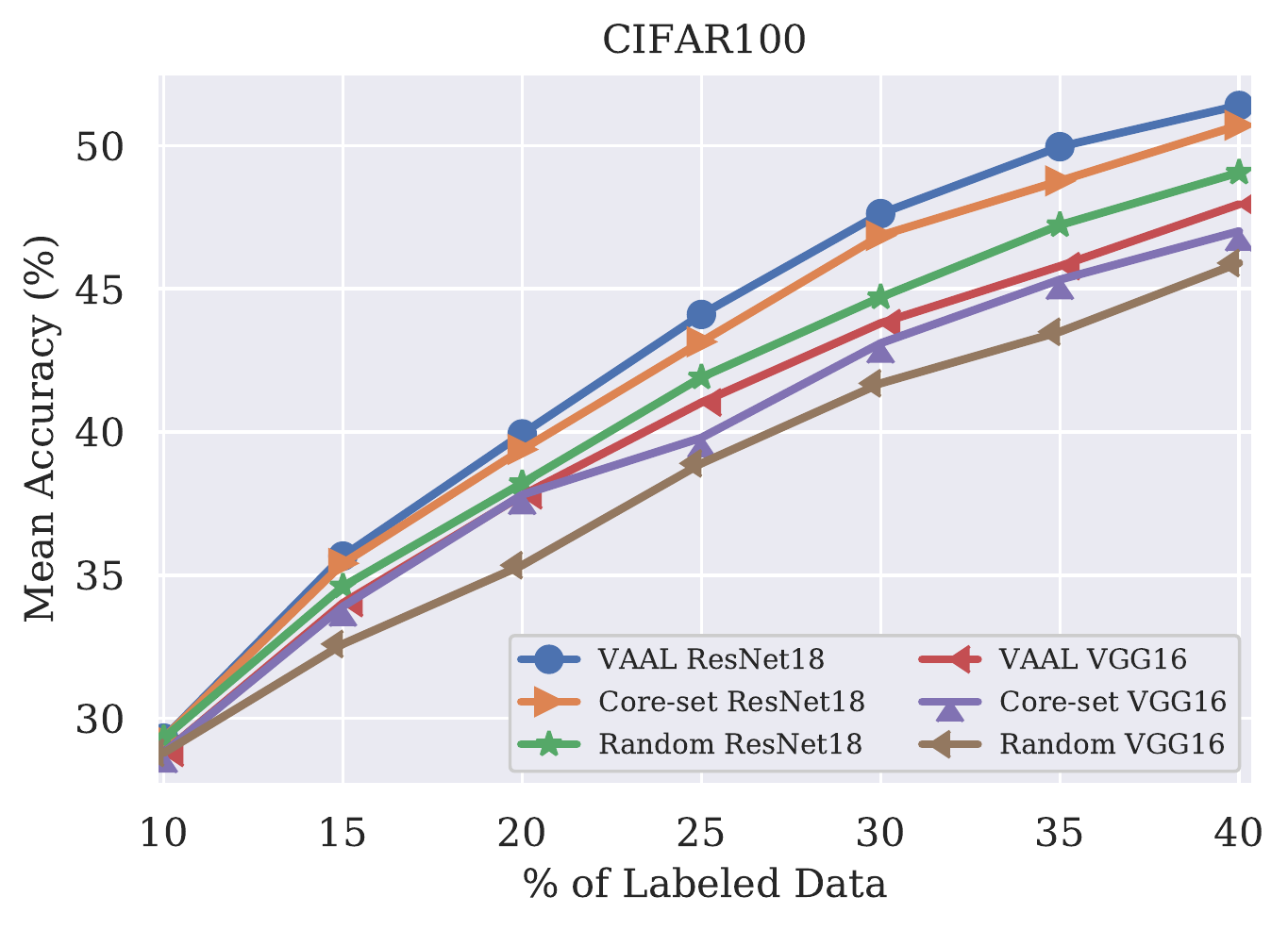}
    \end{center}
    \caption{Performance of VAAL using ResNet18 and VGG16 on $\text{CIFAR100}$. Best visible in color. Data and code required to reproduce are provided in our code repository}
    \label{fig:arch}
\end{figure}

\noindent Figure \ref{fig:arch} shows the performance of our method is  robust to the choice of the architecture by having consistently better performance over Core-set \cite{sener2018coreset} on $\text{CIFAR100}$.

%------------------------------------------------------------------------
%------------------------------------------------------------------------

%------------------------------------------------------------------------
%------------------------------------------------------------------------
%------------------------------------------------------------------------
%------------------------------------------------------------------------
%------------------------------------------------------------------------\\

\end{document}